# From Self-Normal-Positioning to Omni-Directional Tracking: Real-Time Surface Modeling Enabled Probe Tilt Control for Robotic Ultrasound Imaging

Xihan Ma, Haichong K. Zhang

***Abstract*—Ultrasound (US) provides real-time, radiation-free imaging, but the image quality depends strongly on how the probe is oriented against the patient's body. Robotic US can reduce operator workload and improve acquisition consistency; however, most existing systems focus on normal positioning, where the probe is maintained perpendicular to the local surface. This constraint is inadequate for examinations like echocardiography, where obtaining a diagnostic view requires a non-normal probe angle. Consequently, a clinically useful robotic system must sense the local surface in real-time and preserve the desired probe orientation. Here, we propose an omni-directional probe-orientation control framework that integrates RGB-D perception, local-surface modeling, and task-space orientation control. The surface model fuses multi-view point clouds and provides a quadratic estimate of the local surface. A desired imaging direction is then encoded relative to the normal, enabling the probe to track arbitrary angles. The framework was evaluated through flat-surface tracking, phantom target-angle recovery, and in-vivo tracking of an expert-selected view. Results show that the mean angular tracking error was 1.06 ± 0.66°. The system recovered a non-normal tilt angle of up to 44.39 ± 2.59° relative to the surface normal, and acquired the desired heart chamber view in the phantom and in-vivo experiments.**

## I. INTRODUCTION

Medical ultrasound (US) imaging is widely applied in the screening of various anatomies, the diagnosis of common diseases, as well as guiding interventional procedures [1], [2], [3]. It presents unique advantages such as real-time, radiation-free, and relatively low cost [2], [4]. Yet, traditional freehand US procedures remain highly operator-dependent: the image quality and diagnostic reliability are sensitive to probe positioning, contact consistency, and operator skill [4], [5], [6]. Moreover, repeated scanning imposes substantial workload on sonographers, leading to musculoskeletal disorders [7]. To address these issues, robotic ultrasound (RUS) systems have been extensively studied over the past decade. RUS systems employ a robot arm to operate the US probe, thereby alleviating the physical burden to sonographers. Moreover, US probe maneuvers can be automated via accurate robot arm motion, hence improving outcome consistency by standardizing the image acquisition procedure [4], [5], [6], [8].

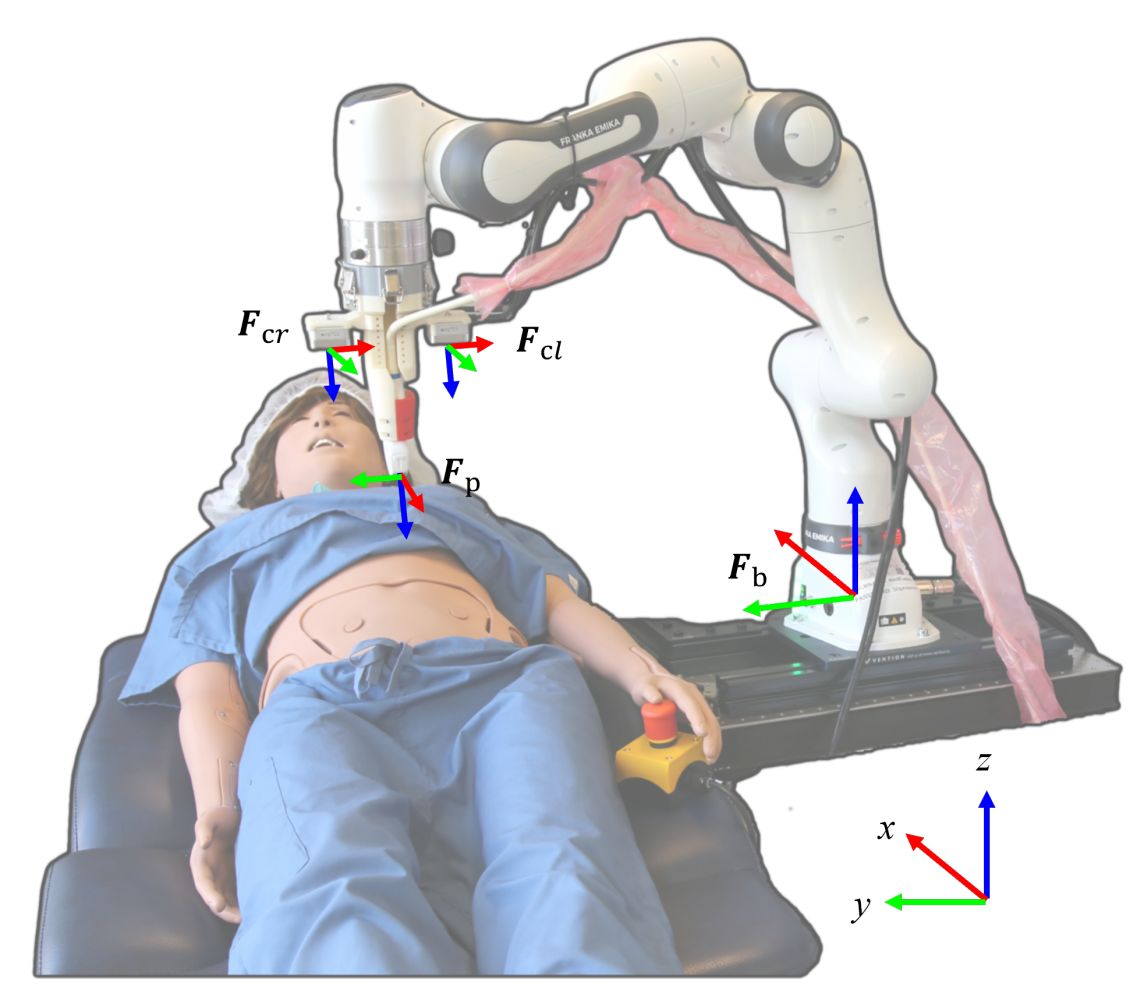

Fig. 1. Robotic US system (RUS) coordinate frame convention. $\{F_b\}$ is the robot base coordinate frame. $\{F_p\}$ is the US probe coordinate frame. $\{F_{cl}\}$ and $\{F_{cr}\}$ are the left and right RGB-D camera frame, respectively.

A critical component in an autonomous RUS system is the probe orientation control mechanism [9], [10], [11]. This is because a slight change in the probe tilt angle can significantly alter both the US signal quality and the anatomies visible in the image. As illustrated by Fig. 2, clinically useful ultrasound views may require the probe to maintain a task-specific non-normal angle with respect to the local body surface. For example, echocardiographic acquisition of the subcostal four-chamber (S4C) view requires the probe to be tilted toward the xiphoid process while maintaining acoustic coupling through the abdominal surface and liver acoustic window [12]. Incorrect alignment of the probe can lead to poor quality images and inaccurate diagnostic outcomes. Therefore, RUS systems need to accurately sense the local skin surface and control the probe orientation relative to this surface, rather than only aligning the probe with the surface normal.

### *A. Related Works on Probe Orientation Control*

As a key component which enables autonomous RUS systems, real-time probe orientation control strategies have been actively explored over recent years. Tsumura *et al.* presented passive end-effectors using spring-based mechanisms that automatically maintain probe contact and normal positioning in both in-plane and out-of-plane

Xihan Ma and Haichong K. Zhang are with the Department of Robotics Engineering, Worcester Polytechnic Institute, 100 Institute Road, Worcester, MA 01609 USA (e-mail: xma4@wpi.edu; hzhang10@wpi.edu).

directions [13], [14]. Related mechanically adaptive end-effectors have also been developed to maintain probe pose and contact forces [15]. These hardware-based approaches ensure fast response and can accommodate patients with different body habitus. Nonetheless, their mechanical footprint can limit clinical usability. To achieve probe orientation control within a more compact system, software-based methods utilizing sensory feedback have been investigated. Jiang *et al.* [9], [10] and Raina *et al.* [16] presented force- and optimization-based strategies for estimating the surface-normal direction. Yet, force-based normal estimation can be sensitive to noisy contact measurements and patient motion. Graumann *et al.* [17] and Huang *et al.* [18] employed RGB-D cameras to acquire body-surface point clouds and estimate probe poses for alignment and scanning. Ma *et al.* further combined RGB-D perception with learned human-body localization to automate lung-US scanning-target placement [19]. However, external vision can be vulnerable to line-of-sight occlusions and may lack the precision needed to track subtle skin deformation. To overcome these limitations, Ma *et al.* proposed a wrist-mounted proximity-sensor array that estimates the normal of the local skin surface, enabling simultaneous in-plane and out-of-plane probe normal positioning [20]. This concept is referred to as the Active-Sensing End-Effector (A-SEE) approach. A revised version uses dual wrist-mounted, close-range RGB-D cameras to provide dense contact-surface information for improved normal-estimation accuracy [21].

### *B. Contributions*

Despite this progress, two primary research gaps remain. First, while the A-SEE concept improves local sensing capabilities, it uses a naïve nearest-neighbor-based normal-estimation technique, which struggles to maintain accuracy on highly curved or irregular anatomies. Second, most surface-perception-based probe-orientation controllers remain functionally restricted to self-normal-positioning, which enforces perpendicular tilt angles both in-plane and out-of-plane. Although prior studies have investigated out-of-plane surface following [22], image-based navigation toward standardized imaging planes [23], and template-guided non-normal pivoting for volumetric acquisition [24], they do not address continuous tracking of an arbitrary, clinically selected surface-relative tilt while the local surface geometry changes during imaging.

To address these limitations, this work builds upon the latest A-SEE hardware design [21] and extends its functionality from “self-normal-positioning” to omni-directional probe orientation control. This is achieved via a novel surface modeling method combined with a generalized omni-directional tracking framework. The contributions of this paper are:

1) We develop a local body-surface modeling pipeline based on dual RGB-D fusion and quadratic surface fitting. Compared with previous local geometric approximations, this formulation improves normal estimation robustness while maintaining real-time behavior needed for stable robotic probe maneuvering
2) We propose a probe orientation control framework that extends from orthogonal-only to omni-directional probe angle tracking. Our system can track a prescribed target tilt angle with respect to the patient’s body surface while maintaining the desired axial imaging-plane constraint. The desired tilt angle can be defined based on specific clinical needs, allowing the robot to acquire diagnostic US images.
3) We present a staged validation of the proposed framework, including flat-surface tilt tracking, cardiac phantom S4C search-and-recovery, and in-vivo expert-selected S4C tracking during respiration. These experiments verify the perception-control accuracy, workflow functionality, and practical feasibility of omni-directional probe angle tracking for echocardiography.

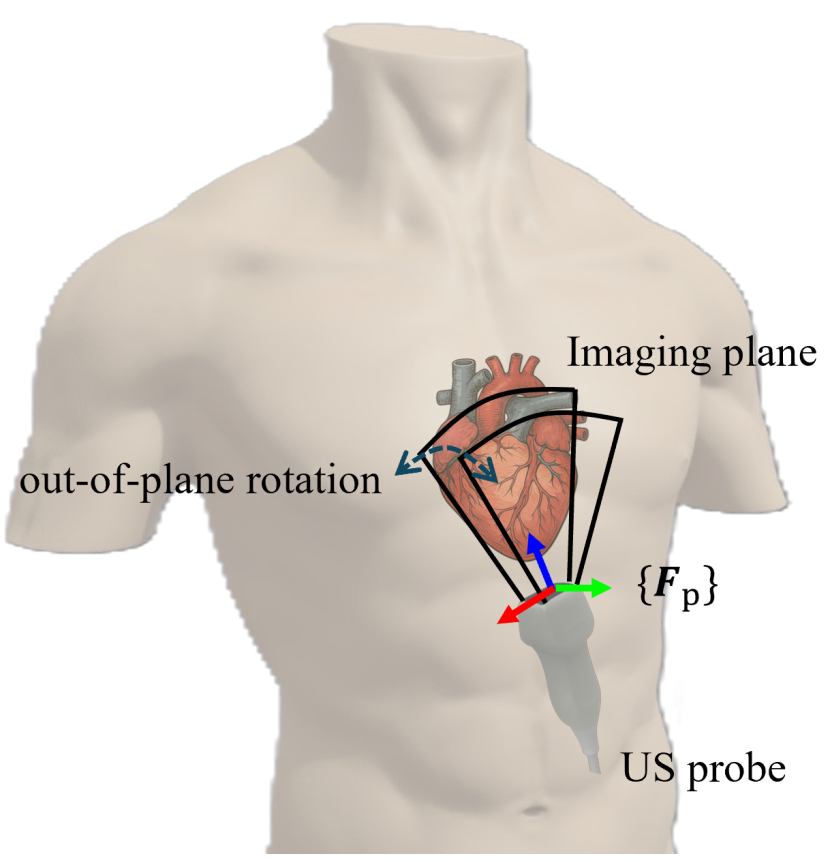

Fig. 2. Illustration of subcostal four chamber (S4C) view imaging. $\{F_p\}$ is the probe coordinate frame.

The rest of this paper is organized as follows: section 2 explains the technical details of the omni-directional tracking framework, section 3 describes the experiments where the proposed framework is validated, section 4 presents the experiment results, section 5 draws the conclusion, discusses limitations and future directions.

## II. MATERIALS AND METHODS

### *A. Problem Statement*

The fundamental objective of this work is to achieve autonomous probe orientation control within a RUS system. As shown in Fig. 1, a local coordinate frame $\{F_p\}$ is defined at the tip of the US probe. The rotation degrees-of-freedom (DoF) of the probe can be decomposed into three distinct components:

- **In-plane Orientation ($\boldsymbol{\theta}_{\mathbf{in}}$)**: Rotation about the y-axis of $\{F_p\}$. This motion adjusts the angle of the beam within the contact-normal plane.
- **Out-of-plane Orientation ($\boldsymbol{\theta}_{\mathbf{out}}$)**: Rotation about the x-axis of $\{F_p\}$. This motion sweeps the imaging plane, changing anatomies visible in the cross-sectional view.

- **Axial Orientation ($\theta_{\text{ax}}$)**: Rotation about the z-axis of $\{F_p\}$. This determines the rotational alignment of the imaging plane itself (e.g., switching between longitudinal and transverse views).

During the US scanning, the robot controller must simultaneously satisfy three hierarchical objectives:

- **Optimal Acoustic Coupling (Adjusting $\theta_{\text{in}}$)**: To ensure maximum contact between the transducer array and the skin, the z-axis of the probe must remain perpendicular to the local skin surface normal $n_s$ within the x-z plane.
- **Task-specific View Alignment (Adjusting $\theta_{\text{out}}$)**: Depending on the clinical protocol, the z-axis must be able to track a target tilt angle α relative to $n_s$ within the x-y plane.
- **Anatomical Plane Consistency (Adjusting $\theta_{\text{ax}}$)**: The axial rotation is predefined according to the desired anatomical plane and held invariant throughout the process of adjusting $\theta_{\text{in}}$ and $\theta_{\text{out}}$.

To satisfy the above control objectives, we define the control problem as the constrained alignment of the end-effector's approach vector with a task-defined target vector $v_t$ in the robot's workspace. The target vector is first defined relative to the estimated local surface normal and is then transformed into the robot base frame for task-space control. Let $R_p^b = [x_{\text{ee}}^{\text{T}}, y_{\text{ee}}^{\text{T}}, z_{\text{ee}}^{\text{T}}] \in \text{SO}(3)$ represent the rotation of the probe with respect to the robot base. The objective is to find the optimal orientation such that:

$$z_{\text{ee}} \cdot v_t \to 1 \tag{1}$$

subject to the following two constraints:

- **Contact Constraint**: $\text{proj}_{\text{xz}}(z_{\text{ee}}) \perp n_s$, ensuring in-plane orthogonality
- **Axial Constraint**: $\dot{\theta}_{\text{ax}} = 0$ , ensuring no rotation about $z_{\text{ee}}$ such that the probe images a fixed anatomical plane

By satisfying these constraints, the RUS system can track a non-normal view direction according to specific diagnostic protocols. In this paper, we choose echocardiography as the example application to demonstrate the clinical usefulness of the proposed omni-directional probe orientation control. Specifically, we demonstrate surface-relative non-normal probe orientation control for S4C imaging, where the probe must be tilted toward the xiphoid process while maintaining contact with the abdominal surface (see Fig. 2).

### *B. Technical Approach Overview*

Fig. 1 shows the RUS system developed for this study. The system comprises of a 7 DoF robot manipulator (FR3, Franka Emika, Germany), a customized end-effector, and a cart-based US machine (Logiq E9, GE Healthcare, USA). The customized end-effector design was first introduced in [21], which encompasses a US probe (C1-6, GE Healthcare, USA) and two RGB-D cameras (RealSense D405, Intel, USA). These RGB-D cameras' nominal depth sensing distance ranges from 7 cm to 50 cm; Therefore, they are ideal for local tissue perception. The spatial transformation from the robot base frame to the probe frame, as well as the transformation from the camera frames to the probe frame, were calibrated in [21].

During imaging, the US probe is firmly pressed on the body surface (section 2.6). The system retrieves filtered point cloud data from both cameras and merges them into a common coordinate frame (section 2.3). A quadratic surface fitting algorithm is employed to reconstruct the local body surface and estimate the surface normal vector (section 2.4). Once the surface normal is obtained, the task-defined target vector $v_t$ is computed based on specific US exam being performed. A probe orientation controller is designed to enable omni-direction probe tilt angle, allowing the system to track tracks $v_t$ in real-time for image acquisition (section 2.5).

### *C. Dual RGB-D Perception Pipeline*

To achieve robust orientation control, the system must accurately reconstruct the local anatomy in real-time. We implemented a perception pipeline, as illustrated in Fig. 3a, to process image data from the dual writ-mounted RGB-D cameras to reconstructed the local contact surface for the subsequent orientation control.

The signal path begins with the acquisition of raw RGB-D streams from both cameras, $C_l$ and $C_r$. To isolate the patient anatomy from the surgical environment, a background masking procedure is applied to the RGB-D data. This is achieved from a combination of depth-truncation and color-thresholding, ensuring that only the tissue foreground representing the local skin manifold is retained. The masked RGB-D data is converted to point clouds using manufacture provided camera intrinsics. Then, the point clouds from each camera, $P^{\text{cl}}$ and $P^{\text{cr}}$, are merged into a unified global point cloud $P^{\text{b}}$ in the robot base frame $\{F_{\text{p}}\}$:

$$P^{\text{b}} = T_p^b \cdot T_{cl}^p (P^{cl} \cup (T_{cr}^{cl} \cdot P^{cr})) \tag{2}$$

where $T_{cl}^p$ and $T_{cr}^{cl}$ are pre-calibrated rigid body transformations determined using a standard checkerboard calibration procedure; $T_p^b$ is the robot's forward kinematics. By utilizing dual perspectives, our pipeline mitigates self-occlusion caused by the probe housing and ensures a dense, multi-view representation of the contact surface.

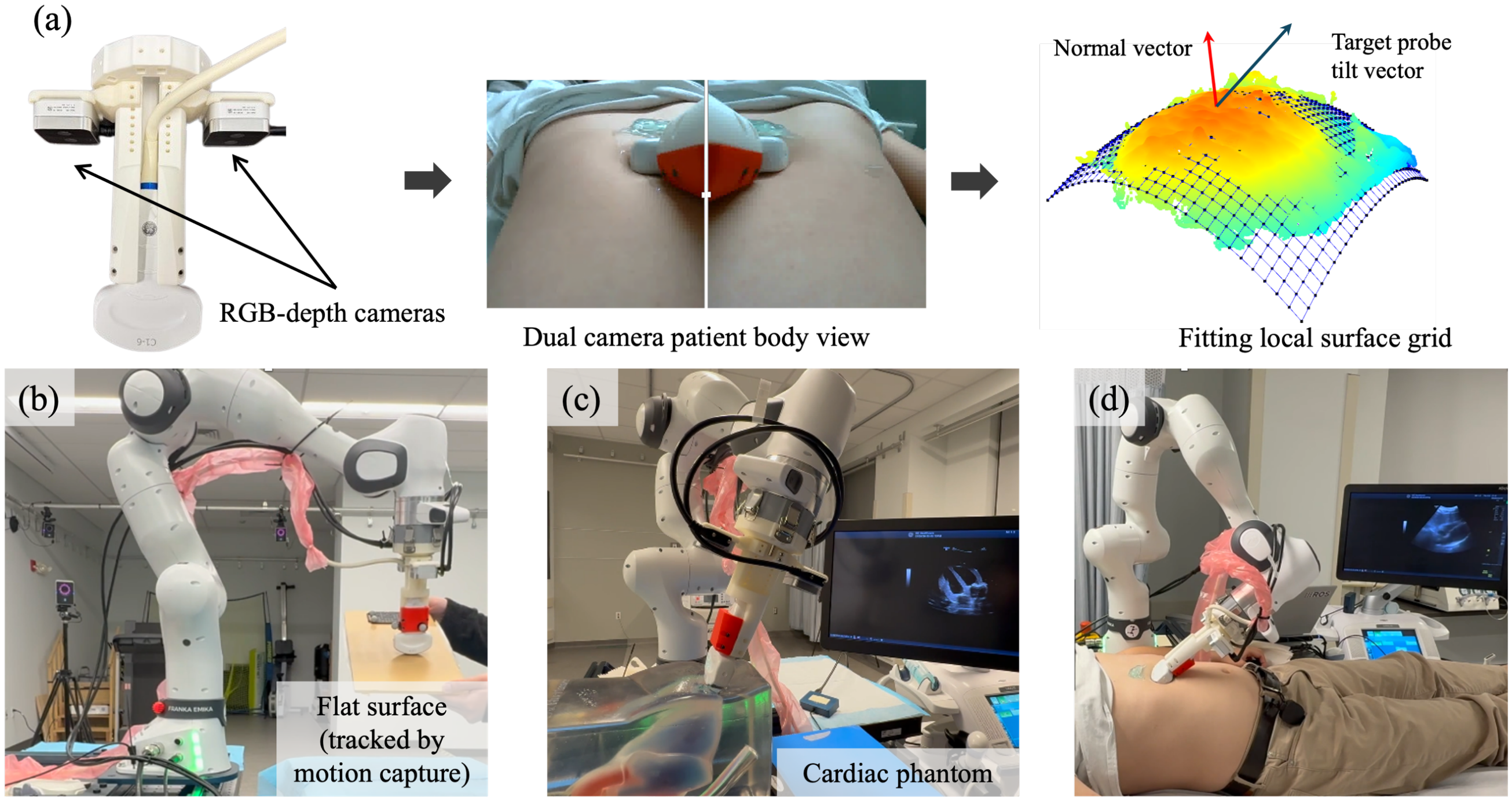

Fig. 3. Omni-directional system workflow and experimental configurations. (a) Dual RGB-D cameras acquire the local body surface, reconstruct the surface point cloud, and estimate the local surface normal and target probe direction for omni-directional tracking. (b) Flat-surface tilt-tracking experiment. (c) Cardiac phantom S4C search-and-recovery experiment. (d) In-vivo S4C tracking experiment..

As the skin surface is often highly curved, we model the local contact manifold using a quadratic surface. This formulation provides a second order approximation of the geometry, capturing essential curvature that a first order model (used in previous works [17], [18], [25]) ignores. We define the surface height $z$ as a function of the lateral coordinates $(x, y)$ in the of the merged point cloud $P^{\mathrm{b}}$:

$$z(x, y) = a_0 + a_1 x + a_2 y + a_3 x^2 + a_4 xy + a_5 y^2 \quad (3)$$

To solve for the coefficient vector $\beta = [a_0, a_1, a_2, a_3, a_4, a_5]^{\mathrm{T}}$, we employ the least squares approach for a set of N points in $P^{\mathrm{b}}$:

$$\beta = (A^{\mathrm{T}} A)^{-1} A^{\mathrm{T}} b \quad (4)$$

where A is the observation matrix defined as

$$A = \begin{bmatrix} 1 & x_1 & y_1 & x_1^2 & x_1 y_1 & y_1^2 \\ \vdots & \vdots & \vdots & \vdots & \vdots & \vdots \\ 1 & x_N & y_N & x_N^2 & x_N y_N & y_N^2 \end{bmatrix}$$

$b = [z_1 \cdots z_N]^{\mathrm{T}}$ is the surface height vector.

The quadratic surface modeling serves two purposes. First, it acts as a spatial low-pass filter, effectively suppressing high frequency sensor noise and outlier present in the raw RGB-D data. Second, it allows for the derivation of a continuous, differentiable surface, enabling the controller to estimate the surface normal much faster than previous nearest-neighbor based methods. The local surface normal $n_s$ at the probe's contact point is derived from the gradient of the implicit surface function $F(x, y, z) = z - f(x, y)$. By taking the partial derivatives of the quadratic equation at the probe tip position $(x_0, y_0)$, we can estimate the normal vector:

$$n_s = \begin{bmatrix} -\dfrac{\partial z}{\partial x} \\ -\dfrac{\partial z}{\partial y} \\ 1 \end{bmatrix} = \begin{bmatrix} -(a_1 + 2a_3 x_0 + a_4 y_0) \\ -(a_2 + a_4 x_0 + 2a_5 y_0) \\ 1 \end{bmatrix} \quad (5)$$

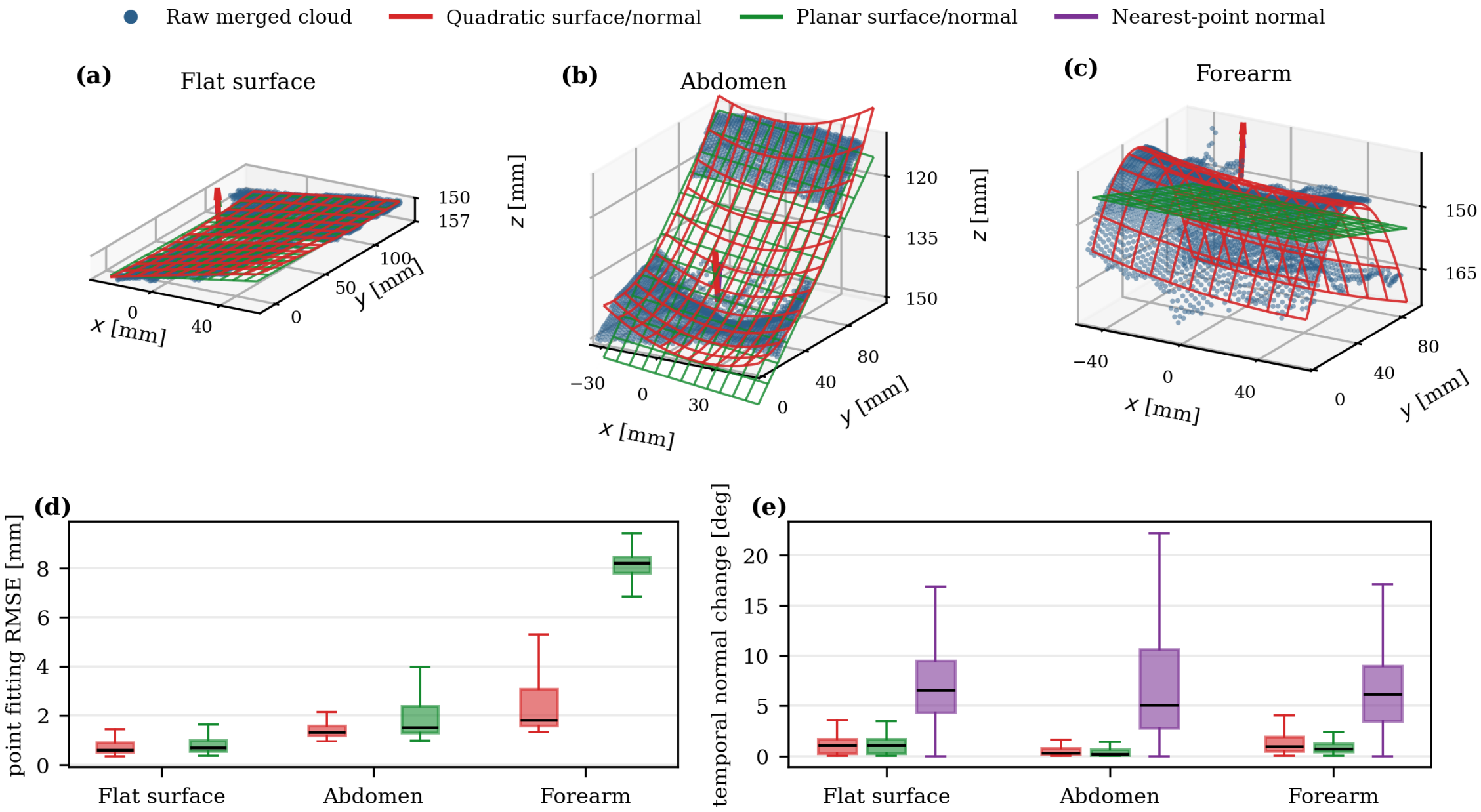

Fig. 4. Comparison of local surface and normal estimation methods. (a)-(c) Representative point clouds from both cameras acquired from a flat, a human abdomen, and a human forearm surfaces, respectively. Blue points are the raw point data. Red and green meshes and arrows show the quadratic and planar fitting and surface normals (d) Point fitting RMSE for the quadratic and planar models. (e) Temporal change in the normal vector estimation when the target surfaces are not moving.

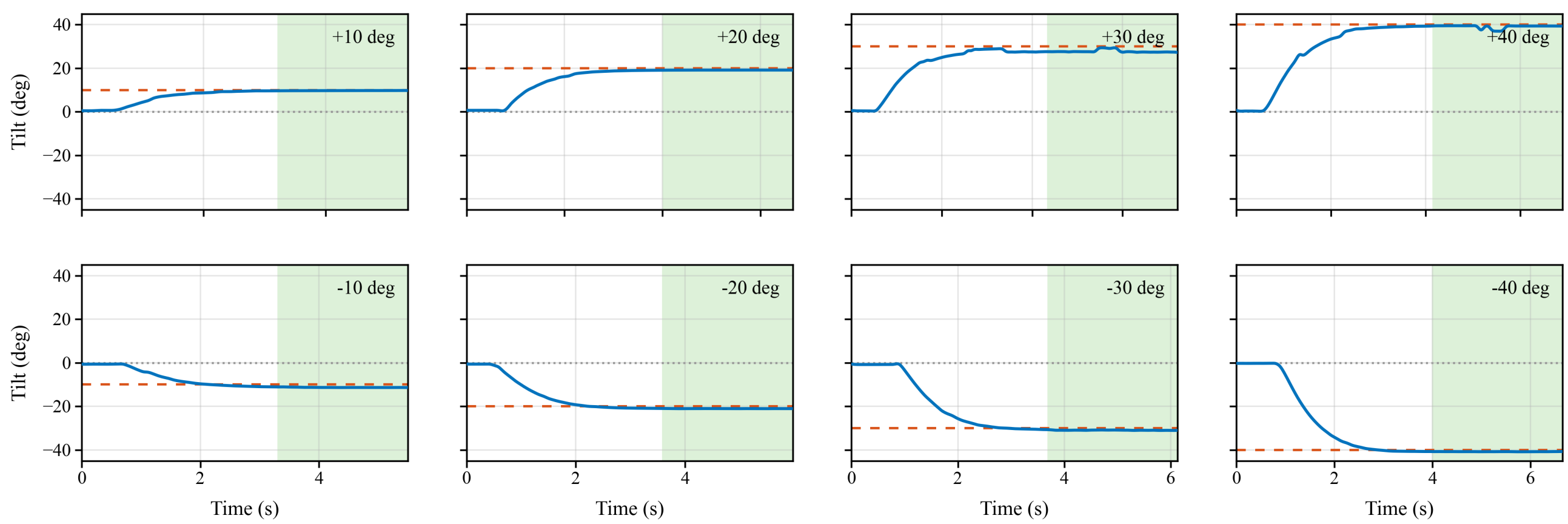

Fig. 5. Measured probe tilt responses for eight commanded out-of-plane target angles from -40 to 40 degrees. Blue curves indicate motion-capture measured tilt, orange dashed lines indicate commanded targets, and green shaded regions indicate the steady-state analysis windows.

Finally, the vector is normalized to unit length. This normal vector serves as the reference “zero” for the subsequent out-of-plane angular adjustments.

To characterize the effect of surface representation on fitting accuracy and normal stability, the same dual-camera point-cloud sequences acquired from a flat surface, a human abdomen, and a human forearm were processed using the proposed quadratic model, a planar model, and the probe-tip-to-nearest-point method (Fig. 4). Surface models were compared using point-fitting RMSE, while temporal normal stability was quantified by the frame-to-frame angular change in the estimated normal.

The quadratic model reduced the median point-fitting RMSE relative to the planar model by 13.2%, 11.8%, and 77.9% for the flat-surface, abdomen, and forearm sequences, respectively. Its median frame-to-frame normal change ranged from 0.32° to 1.09°, compared with 5.09°–6.56° for nearest-point estimation. Although the planar estimates were slightly smoother for the abdomen and forearm, their larger fitting residuals, particularly over the forearm, demonstrate the limitations of a first-order representation on curved anatomy. The quadratic model therefore provides a practical compromise between geometric fidelity and temporal stability. When deploying this perception pipeline on the system, the average point-cloud update rate was between 26.35 to 29.49 Hz, depending on the surface condition. The computation speed of the quadratic surface fitting algorithm was measured to be 0.100 to 0.123 ms per frame, demonstrating sufficient runtime performance.

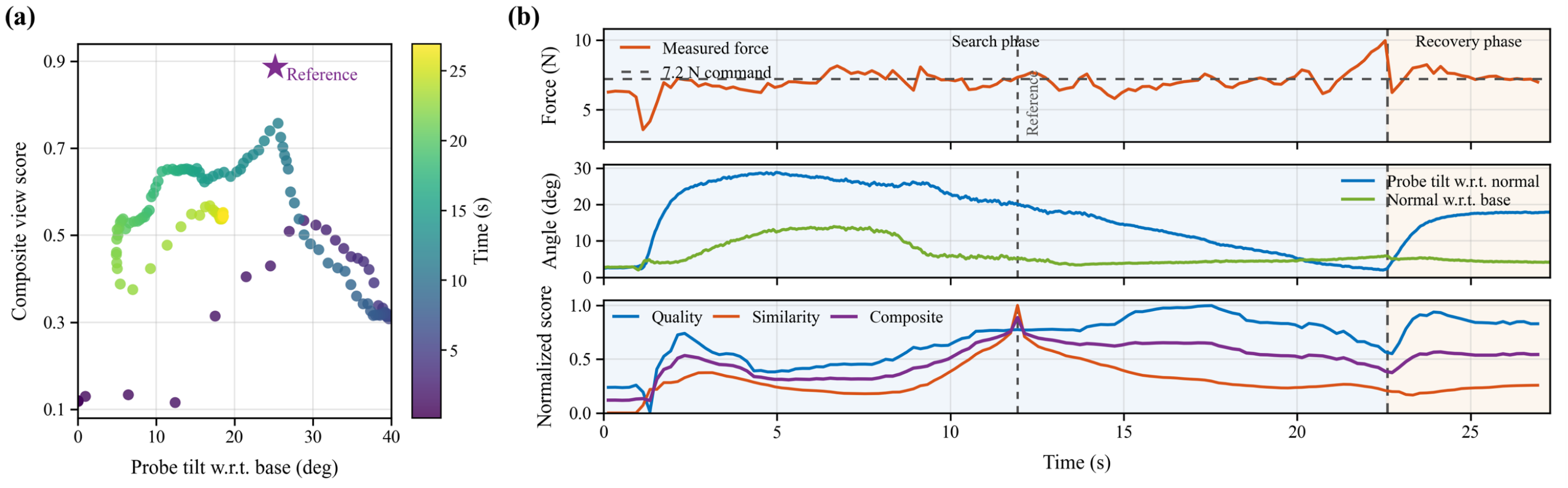

Fig. 6. Cardiac phantom search-and-recovery workflow. (a) Composite view score versus probe tilt during the search, with color indicating time and the star marking the selected reference view. (b) Synchronized force, surface-normal-relative tilt, normal direction with respect to the robot base, and B-mode view-quality metrics during search and recovery.

### D. Omni-Directional Tilt Angle Tracking

The acquisition of the S4C view in echocardiography requires the probe to be placed slightly inferior to the sternum along the body midline. The probe is tilted to a non-normal angular offsets relative to the body surface (see Fig. 2). In this section, we propose a two-stage approach to first identify the proper probe angle, then track this direction dynamically as the probe moves across the anatomy. Ultrasound-image-driven probe optimization and pivoting have been explored previously [11], [23], [24]; the present work instead retains the selected direction relative to the continuously updated local surface.

To identify the optimal probe angle, we utilize an intensity-driven sweeping strategy: First, the probe is vertically placed near the xiphoid process with an axial orientation $\theta_{\mathrm{ax}}^{*}$ aligning the transducer array for a transverse body view. The patient is asked to hold breath and maintain a fixed body posture. The local contact surface is reconstructed using the surface fitting method described in the previous section, and the surface normal vector $n_{\mathrm{s}}$ is obtained. Next, a contact force control is activated to press the probe firmly on the body (will be elaborated in section 2.5). The probe performs an out-of-plane sweep, pivoting about its x-axis over a predefined angular range. During the sweep, US images are sampled and an image quality score is computed within a fan-shaped region of interest (ROI) to identify the tilt angle that yields the best S4C-like view:

$$S = \frac{1}{\mathrm{ROI}} \sum_{(u,v)\in \mathrm{ROI}} I(u,v) \tag{6}$$

where $I(u,v)$ represents the pixel intensity at coordinate $(u,v)$ of US frame $I$, obtained from angle $\theta_{\mathrm{out}}$. The angle $\theta_{\mathrm{out}}^{*}$ which yields the maximum $S$ is recorded as the desired out-of-plane tilt. Similarly, a secondary in-plane sweep is performed where the probe is rotated about its y-axis to center the cardiac structures and maximize the contact at $\theta_{\mathrm{in}}^{*}$. Through the two pivoting motions, the probe is locked at the desired angle. To effectively track this angle with patient motion or during a different imaging session, we represent this angle as a relative rotation $R_{\mathrm{n}}^{a} \in \mathrm{SO}(3)$ between the surface normal $n_s$ and the probe's approach vector $a_{\mathrm{des}}$ (i.e., z-axis of $\{F_{\mathrm{p}}\}$ with respect to $\{F_{\mathrm{b}}\}$). This rotation is computed using the axis-angle representation with rotation axis $k$ and angle $\theta$:

$$\begin{cases} \theta = \arccos(-n_s \cdot a_{\mathrm{des}}) \\ k = \dfrac{-n_s \times a_{\mathrm{des}}}{\|-n_s \times a_{\mathrm{des}}\|} \end{cases} \tag{7}$$

Using the Rodrigues' formula, we can compute $R_{\mathrm{n}}^{a} = \mathrm{I} + (\sin\theta)[k]_{\times} + (1-\cos\theta)[k]_{\times}^{2}$, where $[\cdot]_{\times}$ is the skew-symmetric matrix of a vector. For subsequent imaging, the desired probe angle is recovered from real-time estimated surface normal $n_s(t)$ without repeating the pivoting procedure:

$$a_{\mathrm{des}}(t) = R_{\mathrm{n}}^{a} \cdot (-n_s(t)) \tag{8}$$

We further design a task-space controller to track the desired probe angle. The control objective is to align the current approach vector of the probe $a_{\mathrm{cur}}$ with $a_{\mathrm{des}}$ while simultaneously maintaining the default axial orientation $\theta_{\mathrm{ax}}^{*}$. To achieve this, we propose the following control law:

$$\omega_{\mathrm{p}} = K_{p1}(a_{\mathrm{cur}} \times a_{\mathrm{des}}) + K_{p2}(e_{ax} \cdot a_{\mathrm{cur}})a_{\mathrm{cur}} \tag{9}$$

where $\omega_{\mathrm{p}} \in \mathbb{R}^3$ is the desired probe angular velocity expressed in $\{F_{\mathrm{b}}\}$; $K_{p1}$ and $K_{p2}$ are constant control gains; $e_{ax} = x_{\mathrm{cur}} \times x_{\mathrm{des}}$ computes the probe axial orientation error. In this formulation, $K_{p1}$ regulates the primary task of aligning the probe's approach vector, whereas $K_{p2}$ regulates the secondary task of maintaining the transverse body view. As a result, the probe remains locked relative to the local body, regardless of the patient's global positioning or physiological motion.

*E. Full Arm Control*

This section describes the full arm control pipeline. In section 2.4, we derived the desired probe angular velocity to realize orientation tracking. Here, we further derive the desired linear velocity to achieve automatic contact force regulation and positioning fine-adjustment. Finally, the full task-space command is translated into joint-space commend for robot controlling.

To maintain stable acoustic coupling and ensure patient safety, the system regulates the contact force $F_z$ along the probe's approach axis. We utilize a velocity-based PD controller to minimize the error $e_f = F_{\text{des}} - F_z(t)$ between the desired contact force $F_{\text{des}}$ (empirically determined constant) and the measured force $F_{\text{z}}$ (estimated from joint torques) with respect to the probe frame $\{F_{\text{p}}\}$:

$$v_f = K_p e_f(t) + K_d \frac{de_f(t)}{dt} \tag{10}$$

where $K_p$ and $K_d$ are control gains. This velocity can be translated to the robot base frame via $v_f^{\text{b}} = v_f \cdot a_{\text{cur}}$. While the orientation is handled autonomously, a human operator retains control over the probe's lateral positioning via a joystick controller (SpaceMouse, 3D Connexion, USA). This provides linear velocity inputs $[v_{\text{tel},x}^b, v_{\text{tel},y}^b]$ with respect to the robot base. This shared control paradigm facilitates fine-adjustment of the probe tip to make sure the heart anatomy is visible. The total linear velocity command is thus $v_{\text{p}} = [v_{\text{tel},x}^b, v_{\text{tel},y}^b, v_f^b]$.

Given the complete task-space twist $\xi = [v_{\text{p}}^{\text{T}}, \omega_{\text{p}}^{\text{T}}]^{\text{T}}$, we can compute the required joint velocities as $\dot{q} = J(q)^{\dagger}\xi$, where $J(q)^{\dagger}$ is the Moore-Penrose pseudo inverse of the robot's Jacobian matrix. To ensure operational safety, the resulting velocity commands are passed through a saturation filter that enforces maximum joint speed limits.

## III. EXPERIMENT SETUP

The omni-directional control framework in Section 2 is evaluated at three levels with increasing clinical relevance. First, a flat-surface experiment tested whether the perception-control pipeline can command and maintain prescribed non-normal probe tilt angles relative to a known surface reference (Section 3.1). Second, a cardiac phantom experiment tested the proposed S4C workflow: sweeping the probe over an angular range, selecting a view-compatible target orientation, and recovering that orientation relative to the estimated surface normal after the search (Section 3.2). Third, an in-vivo feasibility experiment tested whether the system can retain an expert-selected S4C probe angle while the abdominal surface moves during spontaneous respiration (Section 3.3). Fig. 3b-d summarizes the experimental configurations.

*A. Evaluate Omni-Directional Tilting Accuracy on a Flat Surface*

The flat-surface experiment evaluated surface-relative tilt tracking accuracy under a controlled geometric condition (Fig. 3b). The probe contacted a rigid planar plate, and motion-capture rigid bodies attached to the plate and the robot/probe assembly provided an external measurement of the angle between the probe axis and the plate normal. Starting from a near-normal configuration, the controller was commanded to hold eight static out-of-plane tilt targets: -40, -30, -20, -10, 10, 20, 30, and 40 degrees. These offsets cover both small angular corrections and the large non-normal rotations required for oblique cardiac views with a curvilinear probe.

For each trial, the motion-capture angle was compared with the commanded target to compute the steady-state mean tilt, absolute error, root-mean-square error, steady-state standard deviation, and settling time. A 2-degree band around the commanded angle was used to define settling. The corresponding tilt time histories are shown in Fig. 5.

*B. Evaluate Automatic Target Angle Selection on a Phantom*

The cardiac phantom experiment evaluated whether the system can execute the image-guided search-and-recovery workflow under repeatable imaging conditions (Fig. 3c). The probe was positioned over a subcostal acoustic window of the cardiac phantom, and the contact-force controller regulated the probe-frame normal force to 7.2 N. The robot then performed an out-of-plane pivot search over a 30-degree range with 1-degree increments. At each increment, the B-mode image was evaluated within a fixed fan-shaped ROI to identify the probe tilt that produced the clearest S4C-like phantom view. The score-versus-tilt trajectory is shown in Fig. 6a.

The image score was designed as a task-specific descriptor specifically for this phantom experiment. It combined an intensity-separation term and a boundary-sharpness term:

$$S_{\text{img}} = \alpha S_{\text{contrast}} + (1 - \alpha) S_{\text{boundary}} \tag{11}$$

Here, $S_{\text{contrast}}$ is the normalized Otsu threshold. $S_{\text{boundary}}$ is mean Sobel-gradient magnitude along the boundary of the dark chamber-like region. The dark region was defined by pixels below 80% of the Otsu threshold. The score is used only to identify a reproducible phantom pose; it is not proposed as a clinically validated ultrasound image-quality measure. The tilt angle that maximized this score was stored as the reference target orientation. After the search, the robot entered a recovery phase and attempted to lock the probe back to this target angle relative to the estimated local surface normal.

The primary tracking variable was the probe tilt with respect to the estimated normal, because this quantity defines the desired imaging orientation in the proposed controller. The normal direction with respect to the robot base was also recorded to show whether the surface reference changed in the base frame; maintaining a fixed base-frame probe orientation would not demonstrate surface-relative tracking. Contact force was recorded to verify that angle recovery did not compromise acoustic coupling. Representative B-mode frames were retained to compare the reference, initial, middle, and final views (Fig. 8a), while the synchronized force, orientation, and image-score traces are shown in Fig. 6b. Success was defined as completion of the search, recovery of a non-normal orientation close to the selected reference, and maintenance of regulated contact force.

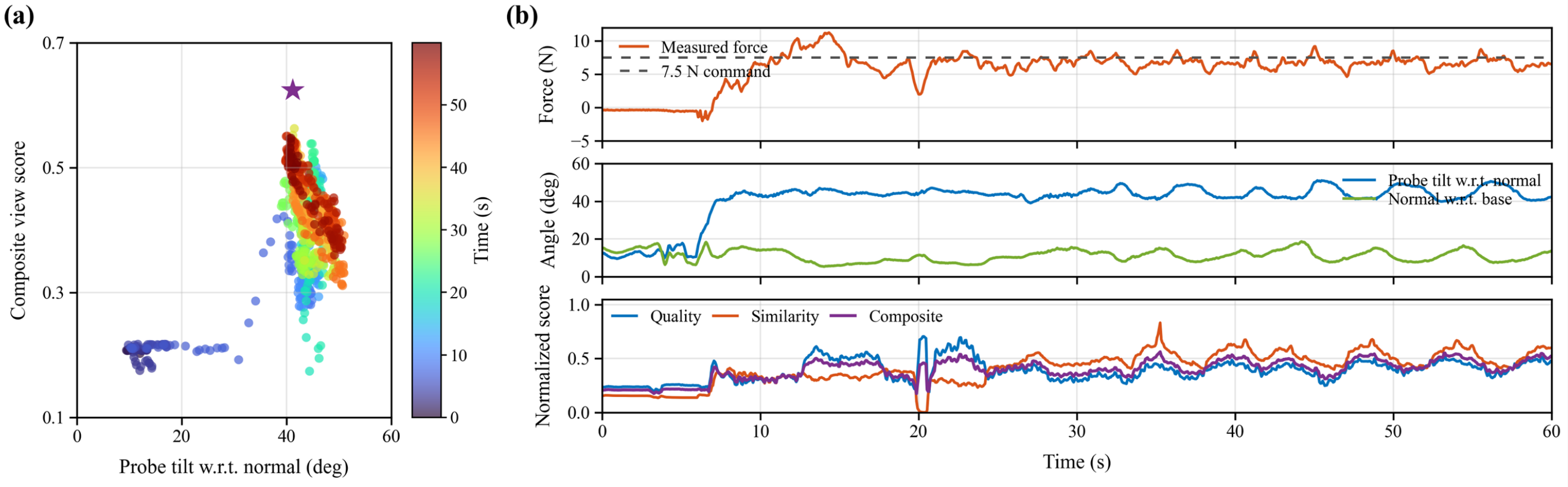

Fig. 7. In-vivo target-angle tracking during respiration. (a) Composite view score versus probe tilt with respect to the estimated surface normal, with color indicating time and the star marking the expert-selected reference view. (b) Synchronized contact force, normal-relative probe tilt, normal direction with respect to the robot base, and B-mode view-quality metrics over the 60-second recording.

### *C. Evaluate Target-Angle Tracking During Respiration In Vivo*

The in-vivo experiment evaluated whether the system can retain an expert-selected non-normal S4C orientation during respiratory motion in a realistic imaging setting (Fig. 3d). The feasibility study involved one participant and was conducted under an institutional review board-approved protocol with written informed consent. Before automated tracking, a human expert manually guided the robot arm to obtain a satisfactory subcostal S4C view with the C1-6 probe. At this pose, the system recorded the probe tilt with respect to the estimated local surface normal, and this expert-defined normal-relative angle was used as the target for automated tracking.

No online image-quality search was performed in vivo. The subcostal acoustic path includes the liver as an acoustic window, and S4C visibility depends on probe position, contact pressure, patient-specific anatomy, and respiration. Therefore, this experiment was designed to evaluate practical target-angle retention rather than automatic clinical image-quality assessment. Once imaging began, the robot directly rotated to the expert-defined angle and continuously regenerated the desired base-frame probe orientation from the real-time estimated surface normal while regulating contact force to 7.5 N.

Data from the first 60 seconds were analyzed. Probe tilt with respect to the estimated normal was the primary outcome because it quantified preservation of the expert-defined imaging angle. The normal direction with respect to the robot base quantified respiratory changes in the surface reference; a changing base-frame normal together with a retained probe-to-normal tilt demonstrates adaptation to body-surface motion. Contact force was analyzed to verify sustained acoustic coupling, and the B-mode stream was reviewed for continuity of the expert-selected S4C view. The score-orientation scatter and synchronized tracking variables are shown in Fig. 7a and Fig. 7b, respectively, while representative B-mode frames are shown in Fig. 8b. Because no external ground-truth normal is available on the deforming abdomen and this study included one participant, success was defined as feasibility: retention of the expert-defined normal-relative tilt, stable contact, and preservation of a recognizable S4C view during respiration.

## IV. RESULTS

### *A. Omni-Directional Tilting Accuracy on a Flat Surface*

Fig. 5 shows the measured tilt responses for the eight commanded targets from -40 to 40 degrees. Across all trials, the steady-state mean absolute error was 1.06 ± 0.66 degrees, and the maximum steady-state standard deviation was 0.13 degrees, indicating that the commanded tilt was held with minimal drift after the transient response. Seven of the eight targets entered and remained within the predefined 2-degree settling band, with settling times from 1.57 to 5.66 seconds. The +30 degree target did not satisfy the settling criterion and reached a steady-state angle of 27.51 degrees, corresponding to a 2.49 degree absolute error. These results verify bidirectional surface-relative tilting over the tested range, with one target exceeding the selected 2-degree engineering threshold.

### *B. Automatic Target-Angle Search and Recovery on a Phantom*

The phantom experiment evaluated whether the robot could search over candidate non-normal orientations, select a view-compatible target, and recover that target after the search. As shown in Fig. 6a, the selected reference view occurred at 11.94 s, corresponding to frame 60 and a probe tilt of 25.19 degrees with respect to the robot base. The reference B-mode frame showed the expected four-chamber phantom anatomy (Fig. 8a), with a composite view score of 0.89.

The search-to-recovery transition occurred at 22.6 s (Fig. 6b). During recovery, the probe returned from the near-normal configuration to an approximately 18-degree tilt with respect to the estimated normal by the final frame at 26.94 s, close to the approximately 20-degree normal-relative tilt at the selected reference. Contact was maintained throughout the analyzed interval, with a mean force of 7.34 ± 0.61 N against the 7.2 N command. The final B-mode frame retained the S4C-like phantom view, with a composite score of 0.54 (Fig. 8a). These results verify that the system completed the search-and-recovery workflow while maintaining acoustic coupling.

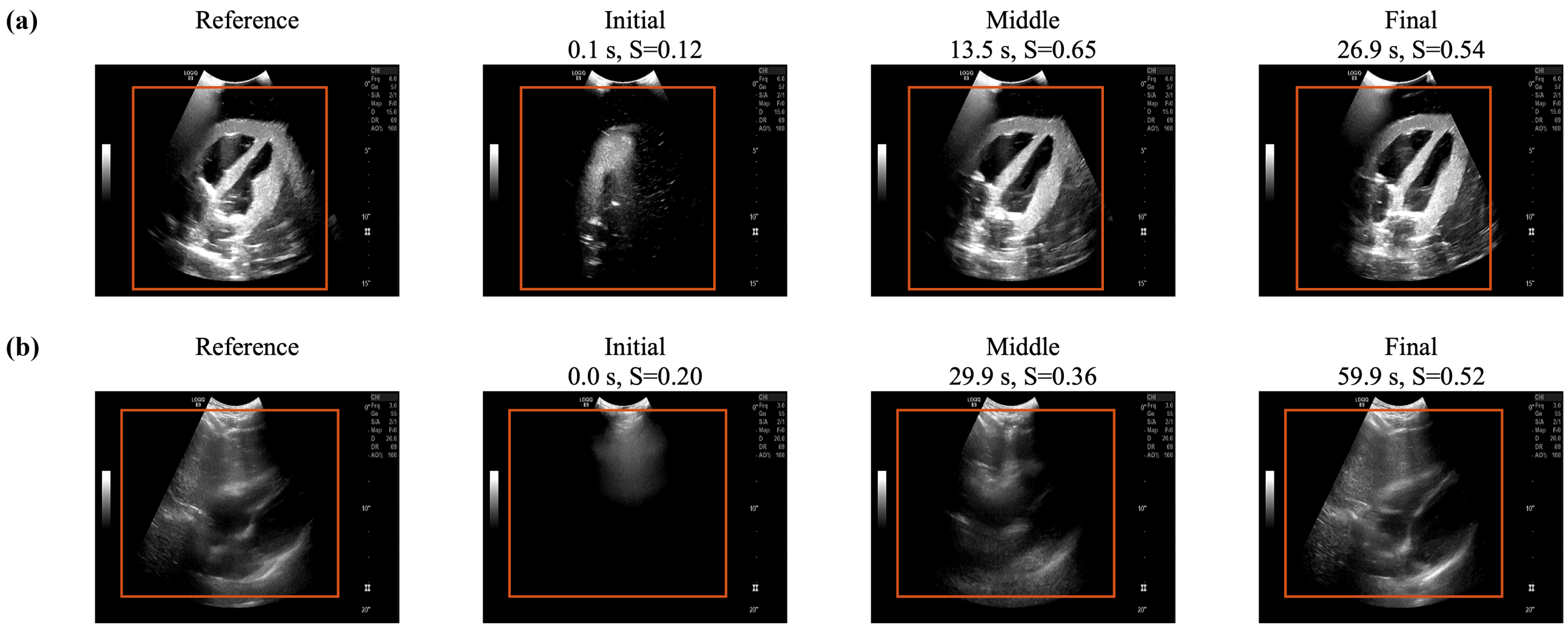

Fig. 8. Representative B-mode image sequences from the cardiac phantom and in-vivo experiments. (a) Cardiac phantom target angle search and recovery. (b) In-vivo target angle tracking during respiration. Each row shows the reference, initial, middle, and final views from left to right. Timestamps and composite view scores $S$ are shown above each frame. Orange boxes delineate the image analysis ROI.

### *C. Target-Angle Tracking During Respiration In Vivo*

The in-vivo experiment evaluated whether the robot could retain an expert-selected S4C orientation while the abdominal surface moved during respiration. The 60-second recording included 900 B-mode frames and 21,543 synchronized pose and force samples. The expert-selected reference at 35.20 s had a probe-to-normal tilt of 41.20 degrees. Across the full 60-second window, the measured probe-to-normal tilt was 40.82 ± 10.12 degrees, including the initial rotation into the target orientation. After the initial transition at approximately 8 s, the normal-relative tilt remained between 39.02 and 51.00 degrees, with a mean of 44.39 ± 2.59 degrees. The final frame at 59.94 s had a normal-relative tilt of 42.19 degrees (Fig. 7b), and the corresponding B-mode image is shown in Fig. 8b.

The normal direction with respect to the robot base varied cyclically during the recording, while the probe-to-normal tilt remained within a narrower tracking band (Fig. 7b). This result indicates that the robot updated its absolute orientation in response to surface motion rather than holding a fixed base-frame pose. The full-window contact-force mean was 6.42 ± 1.98 N, and the B-mode stream was acquired continuously after contact was established. The representative frames show preservation of the subcostal S4C view from the expert-selected reference to the final frame, with the composite descriptor increasing from 0.20 at the initial frame to 0.52 at the final frame (Fig. 8b). These results support the feasibility of retaining an expert-defined normal-relative probe angle during in-vivo respiratory motion.

## V. CONCLUSION AND DISCUSSION

In this work, we developed an omni-directional probe orientation control framework for robotic US imaging. The central problem addressed is that clinically useful US views often require the probe to be tilted to a task-specific non-normal angle relative to the body surface, rather than simply aligned with the surface normal. This capability is particularly important for echocardiography, where acquisition of the subcostal four-chamber view requires the probe to maintain an oblique acoustic window while the local body surface may move with respiration. By combining dual RGB-D surface perception, local surface modeling, normal-relative target-angle representation, and task-space force/orientation control, the proposed system enables the robot to lock the probe to a desired surface-relative direction during imaging.

The flat-surface experiment showed that the robot can command and hold bidirectional non-normal tilt angles over the tested range, with a mean steady-state absolute error of 1.06 degrees and seven of eight targets satisfying the 2-degree settling criterion. The cardiac phantom experiment showed that the robot can sweep over candidate orientations, select an S4C-like view, and recover a similar normal-relative orientation while maintaining acoustic coupling; the corresponding B-mode sequence is shown in Fig. 8a. The in-vivo feasibility experiment further showed that an expert-selected S4C probe angle can be retained during respiratory surface motion in one consented participant, with representative view progression shown in Fig. 8b. Together, these results validate the core functionality required by the proposed framework: the system can estimate the local surface reference, express a clinically motivated probe angle relative to that reference, and update the robot orientation to preserve this angle during imaging.

The results also highlight an important design insight for robotic US systems: The clinically relevant quantity is not the probe orientation in the robot base frame, but the probe angle relative to the local body surface. In both phantom and in-vivo experiments, the surface reference changed over time, while the robot regenerated the corresponding base-frame orientation from the estimated normal. This supports the use of normal-relative target representation as a practical bridge between clinical view selection and robotic control.

This work has several limitations. First, the in-vivo study was conducted on one participant and should be interpreted as feasibility validation rather than population-level clinical validation. Second, the phantom image score was used only to obtain a reproducible S4C-like target in a controlled setting; it was not intended to establish a general clinical image-quality assessment algorithm. Third, the in-vivo experiment used an expert-selected target angle rather than online autonomous image-quality optimization, because the subcostal acoustic window depends on patient-specific anatomy, liver coupling, contact pressure, and respiration. Finally, the current evaluation focused on probe orientation and contact maintenance, while full clinical deployment will also require robust autonomous positioning and complete examination workflow integration.

Future work will extend the evaluation to larger in-vivo cohorts, broader ultrasound views, and more diverse body-surface geometries. Further development should also integrate more reliable clinical image feedback, improve coupled position-orientation control, and evaluate the system in longer imaging workflows. These directions will help translate omni-directional surface-relative probe control from technical feasibility toward clinically deployable robotic US imaging.

## ACKNOWLEDGMENT

The authors declare no conflicts of interest.